\pdfoutput=1
\documentclass{article}
\usepackage{spconf,amsmath,graphicx}
\usepackage{times}
\usepackage{epsfig}
\usepackage{amssymb}
\usepackage{subcaption}
\usepackage{enumitem}
\usepackage{booktabs}
\usepackage{multirow}

\usepackage[pagebackref=true,breaklinks=true,colorlinks,bookmarks=false]{hyperref}

\DeclareMathOperator{\sign}{sgn}

\DeclareMathOperator*{\argmin}{arg\,min}

\title{Level Set Stereo for Cooperative Grouping with Occlusion}

\name{Jialiang Wang and Todd Zickler}
\address{Harvard University}
\begin{document}
%
\maketitle
\begin{abstract}
Localizing stereo boundaries is difficult because matching cues are absent in the occluded regions that are adjacent to them. We introduce an energy and level-set optimizer that improves boundaries by encoding the essential geometry of occlusions: The spatial extent of an occlusion must equal the amplitude of the disparity jump that causes it. 
In a collection of figure-ground scenes from Middlebury and Falling Things stereo datasets, the model provides more accurate boundaries than previous occlusion-handling techniques. 

\end{abstract}
\section{Introduction}
\label{sec:intro}


Deep convolutional networks can provide fast and accurate estimates of binocular disparity by exploiting local and non-local patterns of scene shape and appearance~\cite{poggi2020synergies}, but their reliance on spatial sub-sampling (i.e., stride and pooling) limits their accuracy near object boundaries~\cite{wang2019local,middlebury_website}. A natural way to address this is to develop bottom-up systems for disparity estimation that can eventually be combined with the fast, feed-forward estimates from a CNN, and that complement their top-down disparity information by analyzing local disparity signals at high spatial resolution and explicitly modeling the smooth, curvilinear behavior of boundaries.

This paper takes a step in this direction by introducing a bottom-up approach to stereo disparity estimation in a level set framework. A key contribution is an energy function that properly accounts for occlusions at disparity boundaries, where occluding foreground surfaces cause adjacent sets of scene points to be visible in only one of the two input images. 
Occlusions have so-far been ignored in previous formulations of level-set binocular stereo algorithms.

Our approach is preliminary because it can only produce two-layer (foreground/background) disparity maps. Nevertheless, it has some promising benefits. For one, it accurately hard-codes the geometrical constraint that, in a rectified stereo pair, the spatial extent of an occluded region is equal to the magnitude of the disparity jump that causes it~\cite{belhumeur1996bayesian}. This constraint is properly enforced in one-dimensional algorithms that operate on isolated stereo scanlines~\cite{belhumeur1996bayesian,bobick1999large,wang2017toward} but is only approximated in previous two-dimensional algorithms, for example by: estimating per-pixel occlusion labels that are unconstrained by adjacent disparity jumps~\cite{zitnick2000cooperative, kolmogorov2001computing, lin2004surfaces} or that  are softly constrained by the polarity (but not magnitude) of these jumps~\cite{yamaguchi2012continuous}; performing disparity estimation twice from left and right viewpoints and checking their consistency~\cite{fua1993parallel,sun2005symmetric,hirschmuller2007stereo}; or augmenting training images with fake occlusions when training a neural network~\cite{yang2019hierarchical}. Another benefit is that our model's computations can potentially be made fully cooperative with a Chakrabarti hierarchy~\cite{chakrabarti2015low}, making it amenable to future parallelization and to unrolled, end-to-end training in combination with top-down processing. 

We evaluate our model by measuring the accuracy of estimated disparity boundaries in a small collection of synthetic and manually-annotated captured images curated from existing stereo benchmarks. With approximate initialization, our model converges to estimates of foreground disparity boundaries that are more accurate than those of existing techniques.

\section{Energy function and optimization}
\label{sec:method}
We follow~\cite{belhumeur1996bayesian} and represent disparity as a function on the visual field $\Omega\subset\mathbb{R}^2$ of a virtual cyclopean camera that is rectified and centered between the left and right cameras, associating a disparity value $d \in [0,d_{\max}]$ with each ray $(x,y)\in\Omega$. We restrict the disparity function to be piecewise smooth.  Specifically, we define global basis functions $\mathbf{U}(x,y) = \{U_i(x,y)\}_{i=1 \cdots m}$, and within the $j$th smooth piece we restrict the disparity function to a linear combination $\Theta_j(x,y)= \sum_{i=1}^m \mathbf{\Theta}_j(i) U_i(x,y)$ with shape coefficients $\mathbf{\Theta}_j\in\mathbb{R}^m$. We refer to $\mathbf{\Theta}_j$ as \emph{global shapes}. We use $\mathbf{U}(x,y) = \{x^2, xy, y^2, x, y, 1 \}$
and the convention $\mathbf{\Theta}_1$ and $\mathbf{\Theta}_2$ for foreground and background, respectively. 


To obtain foreground/background boundaries, we evolve a continuous level set function $\phi(x,y)$ that is zero-valued at the boundary, positive-valued in the foreground, and negative valued in the background. It evolves in response to three driving forces. The first force derives from a stereo matching cost $M(x,y,d)$. 
The second force comes from a stereo occlusion boundary cost $B_{\text{occ}}(x,y,d)$ that is inversely proportional to the local evidence for scene point $(x,y,d)$ being on an occluding boundary. 
The third force comes from a monocular boundary cost $B_{\text{mono}}(x,y,d)$, which reflects the fact that occlusion boundaries often co-occur with spatial changes of intensity, color or texture in the images.
We combine these into the energy function
\begin{equation}
\small
\medmuskip=0mu 
\thinmuskip=0mu
\thickmuskip=0mu
\begin{split}
    J(& \mathbf{{\Theta}_1}, \mathbf{{\Theta}_2}, x,y, \phi,  \nabla\phi) = \int_\Omega H(\phi)M_{\Theta_1}dxdy  \\ 
&   +\int_\Omega(1-H(\phi_{+})) (1-H(\phi))M_{\Theta_2}dxdy 
    + \mu \int_\Omega B_{\Theta_1} \delta(\phi) |\nabla \phi|dxdy
\end{split}
\label{eqn:loss_function}
\medmuskip=4mu 
\thinmuskip=3mu
\thickmuskip=5mu
\end{equation}
\vspace{-4mm}

\noindent where $H(\cdot), \delta(\cdot)$ are Heaviside and Dirac delta functions, and
\begin{equation*}
\small
\begin{split}
    & \phi = \phi(x,y), \quad \phi_{+} = \phi(x+\Delta\theta(x,y;  \mathbf{\Theta}_1,\mathbf{\Theta}_2),y), \\
    & M_{\Theta_1} = M(x,y,\Theta_1(x,y)), \quad M_{\Theta_2} = M(x,y,\Theta_2(x,y)), \\
    & B_{\Theta_1}=\alpha_1 B_{\text{occ}}(x,y,\Theta_1(x,y)) + \alpha_2 B_{\text{mono}}(x,y,\Theta_1(x,y)) + \alpha_3
\end{split}
\end{equation*}
with tunable parameters $\alpha_i, \mu$. The first term of Eqn.~(\ref{eqn:loss_function}) is easy to interpret  as the integrated matching cost of the foreground surface, and the third is the weighted length of the foreground boundaries with weight $B_{\Theta_1}$. The second term integrates the matching cost of the background surface, \emph{but only over the subset that is not occluded}.
The expression relies on an intermediate function $\Delta\theta(x,y; \mathbf{\Theta}_1,\mathbf{\Theta}_2)$ that is determined 
as follows: for each $(x,y)$, cast a ray in $x$-$d$ space from $(x,\Theta_1(x,y))$ in the direction of decreasing disparity with slope equal to $\sign(\frac{d\phi(x,y)}{dx})$.
If the ray does not intersect $\Theta_2(x,y)$ then set $\Delta\theta(x,y; \mathbf{\Theta}_1,\mathbf{\Theta}_2)=0$, otherwise find the closest intersection $(x',y)$ and set $\Delta\theta(x,y; \mathbf{\Theta}_1,\mathbf{\Theta}_2)= x-x'$. 
The geometric correctness of this formulation follows from~\cite{wang2019local,belhumeur1996bayesian}.

We use alternating updates to find $\mathbf{{\Theta}_1}, \mathbf{{\Theta}_2}$ and $\phi(x,y)$ that locally minimize $J$. Beginning with some initialization $\phi_0(x,y)$ and at first assuming $\Delta\theta(x,y; \mathbf{\Theta}_1,\mathbf{\Theta}_2)=0$, we update global models $\mathbf{{\Theta}_1}$ and $\mathbf{{\Theta}_2}$ by using the current $\phi(x,y)$ and $\Delta\theta(x,y; \mathbf{\Theta}_1,\mathbf{\Theta}_2)$ to define the foreground and visible-background regions, and then separately solving for the optimal shape parameters $\mathbf{{\Theta}_1}$ and $\mathbf{{\Theta}_2}$ in each region using weighted linear least squares. We then update $\Delta\theta(x,y; \mathbf{\Theta}_1,\mathbf{\Theta}_2)$ as described above.
Finally, we update $\phi(x,y)$ by following the common  practice (e.g.,~\cite{caselles1997geodesic, chan2001active}) of replacing $\delta(\cdot)$ and $H(\cdot)$ with differentiable approximations $\delta_\epsilon(\cdot)$ and $H_\epsilon(\cdot)$, and iteratively solving the Euler-Lagrange equation by descent, with the descent parameterized by $t \ge 0$. Assuming the global models are fixed, one derives
\begin{equation}
\small
\label{eqn:dphi_dt}
\medmuskip=0mu 
\thinmuskip=0mu
\thickmuskip=0mu
\begin{split}
        & \frac{d\phi}{dt} = \delta_\epsilon(\phi) \biggl[  -  M(x,y, {\Theta}_1(x,y)) \\  & + M(x-\Delta\theta(x,y;\mathbf{\Theta}_1,\mathbf{\Theta}_2),y,{\Theta}_2(x,y)) \\
        & + \mu \Big( B(x,y,{\Theta}_1(x,y)) \kappa(x,y) +  \mathbf{N}(x,y) \cdot \nabla B(x,y,{\Theta}_1(x,y)) \Big) \biggl], 
\end{split}
\medmuskip=4mu 
\thinmuskip=3mu
\thickmuskip=5mu
\end{equation}

\noindent with  $\frac{\delta_\epsilon(\phi)}{|\nabla \phi|}\frac{d\phi}{d\vec{n}} = 0$ on the boundary $\partial \Omega$ of the visual field.
Here, $\kappa(x,y)=\text{div} (\frac{\nabla \phi(x,y)}{|\nabla \phi(x,y)|})$ and $\mathbf{N}(x,y) = \frac{\nabla \phi(x,y)}{|\nabla \phi(x,y)|}$
are the curvature and normal of the foreground contour, and $\vec{n}$ is the exterior normal to $\partial \Omega$. 
Upon convergence, the final disparity map is 
\begin{equation}
\small
\label{eqn:D}
    D(x,y) = H(\phi(x,y))\Theta_1(x,y) + (1-H(\phi(x,y)))\Theta_2(x,y).
\end{equation}
\vspace{-6mm}

\begin{figure}[t!]
    \centering
    \includegraphics[width=0.48\textwidth]{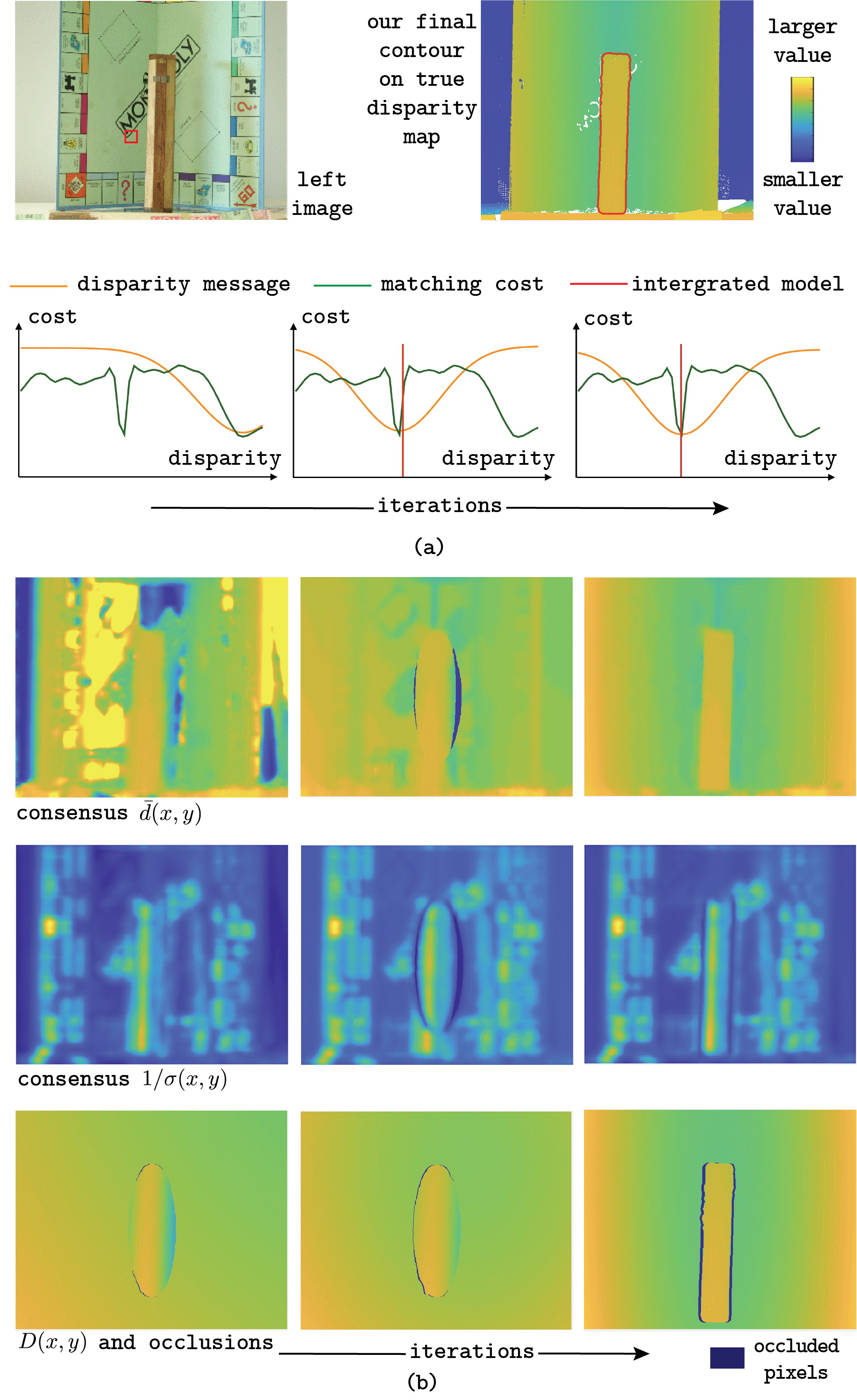}
    \caption{
     (a) Evolution of disparity message for the red highlighted patch, visualized using Eqn.~\ref{eq:patch-gaussian}, along with patch mean matching cost $\langle M(x,y,d)\rangle_{(x,y)\in p}$ and patch mean of evolving two-plane fit $\langle D(x,y)\rangle_{(x,y)\in p}$. 
    The patch is initially erroneous due to erratic behavior of local matching cost (caused by repeated texture), but message corrects over time.
    (b) Evolution of consensus $\left(\bar{d}(x,y),\sigma(x,y)\right)$ and two-plane fit $D(x,y)$ with occlusions shown in dark blue. Consensus mean $\bar{d}(x,y)$ is initially erroneous in many places but expresses high variance $\sigma(x,y)$ at locations of large error, so influence on $D(x,y)$ is appropriately suppressed. Error decreases as the consensus guides the boundary toward the true one.}
    \label{fig:patch_and_pixel_vote}
\end{figure}

\section{Alternating Descent}

It is challenging to implement the alternating approach in a way that succeeds despite the erratic behavior of stereo matching costs $M(x,y,d)$, which are caused by textureless regions, repeated textures, and so on. 
Our strategy is to do matching in dense overlapping patches at multiple scales, and to aggregate their disparity information using an evolving family of per-pixel Gaussian disparity distributions, which we call the \emph{consensus}. The consensus is visualized in Fig.~\ref{fig:patch_and_pixel_vote}(b) by its means $\bar{d}(x,y)$ and inverse standard deviations $1/\sigma(x,y)$. At each iteration, the consensus summarizes disparity information from all unoccluded patches.
The alternation involves: (i) each patch updating its occlusion status (``validity'') and a ``disparity message'' based on the current boundary and global models; (ii) per-patch disparity messages being collected in the consensus; and (iii) global models $\mathbf{\Theta_1},\mathbf{\Theta_2}$ and boundary $\phi(x,y)$ being updated from the consensus.

Specifically, let $\mathcal{P} = \{ p \}$, a set of densely overlapping patches $p$ of multiple sizes, including a complete subset of patches that each comprise a single pixel $(x,y)$. Assume there exists a stereo cost volume at the pixel resolution, $\{M(x,y,d), d \in [0,d_{\max}]\}$, and an initial consensus $\left(\bar{d}(x,y),\sigma(x,y)\right)$, global models $\mathbf{\Theta_1},\mathbf{\Theta_2}$ and boundary $\phi(x,y)$. Equip each patch with an evolving state $\{w_p, d_p,\sigma_p\}$ representing a patch's validity (i.e., occlusion status) $w_p\in\{0,1\}$ and a per-patch disparity message $(d_p,\sigma_p)\in\mathbb{R}^2$.

The three update stages are as follows, with Fig.~\ref{fig:patch_and_pixel_vote} visualizing the dynamics. 


\vspace{1mm}
\noindent \textbf{Updating per-patch disparity messages.}
We initially assume all patches are valid: $\forall p, w_p = 1$. At subsequent iterations, each patch updates its validity using $w_p =$
\begin{equation}
 \begin{split}
\small
  \max \limits_{(x,y) \in p}  &  \phi(x,y)  > 0  \oplus
  \min\limits_{(x,y) \in p} \phi(x+\Delta\theta(x,y;\mathbf{\Theta}_1,\mathbf{\Theta}_2),y)  < 0
   \end{split}
\end{equation}
\noindent where $\oplus$ is the logical XOR operator. This says that a valid patch is neither contained in an occluded regions nor includes portions of both foreground and visible-background.

Each valid patch also updates its disparity message based on a combination of its local matching cost and the current (regularizing) piecewise-smooth global disparity map:
\begin{equation}\label{eq:local_message}
\small
    d_p = \argmin_{d}  C_p(d)  \quad \text{and} \quad \sigma_p = \frac{d_\text{max}}{\langle C_p(d)\rangle- \min{C_p(d)}},
\end{equation}
where $C_p(d) = \sum_{(x,y) \in p} ( M(x,y,d)  + \beta | d - D(x,y)| )$ with $\beta$ a tunable parameter and $D(x,y)$ computed using Eqn.~(\ref{eqn:D}).


Figure~\ref{fig:patch_and_pixel_vote}(a) shows an example of the evolving local disparity messages. We interpret $d_p,\sigma_d$ as parameters of an evolving Gaussian approximation, and depict them by drawing
\begin{equation}\label{eq:patch-gaussian}
\small
    f_p(d)=\max{C_p(d)}-\left(\max{C_p(d)}-\min{C_p(d)}\right) e^{\frac{-(d-d_p)^2}{2\sigma^2_p}}.
\end{equation}
\noindent \textbf{Updating consensus.}
 The consensus aggregates the local disparity messages from all valid patches and is updated using the product of Gaussians:
 \begin{equation}
 \small
 \begin{split}
        \frac{1}{\sigma^2(x,y)} &= \sum_{\substack{p \ni (x,y) \\  w_p =1}} \frac{1}{\sigma_{p}^{2}}, \quad 
      \overline{d}(x,y) =\Biggl[\sum_{\substack{p \ni (x,y) \\  w_p =1}} \frac{d_{p}}{\sigma_{p}^{2}}\Biggl] \sigma^{2}(x,y),
 \end{split}
 \label{eqn:pixel_consensus_vote}
 \end{equation}
The first two rows of Fig.~\ref{fig:patch_and_pixel_vote}(b) show the evolving consensus. The consensus means $\bar{d}(x,y)$ are initially erroneous in significant portions of the scene, but with high variance. The consensus corrects over iterations.

\begin{table*}[t!]
\centering
\small
\setlength{\tabcolsep}{2.7pt}
\begin{tabular}{lcccccccccccccccccc}
\toprule
Image & 1 & 2 & 3 & 4 & 5 & 6 & 7 & 8 & 9 & 10 & 11 & 12 & 13 & 14 & 15 & & Avg occlusion F1 \\
\midrule 
SGM~\cite{hirschmuller2009matching} & 0.39 & 0.38 & 0.42 & 0.65 & 0.23 & 0.32 & 0.00 & 0.02 & 0.58 & 0.67 & 0.04 & 0.87 & 0.27 & 0.61 & 0.72 & & 0.41 \\
BM-LR & 0.37 & 0.34 & 0.48 & 0.70 & 0.01 & 0.22 & 0.00 & 0.00 & 0.57 & 0.72 & 0.18 & 0.92 & 0.07 & 0.59 & 0.44 & & 0.37 \\
KZ~\cite{kolmogorov2001computing} & 0.14 & 0.55 & 0.76 & \textbf{0.78} & 0.76 & 0.53 & \textbf{0.68} & 0.70 & \textbf{0.83} & 0.82 & 0.57 & 0.90 & 0.46 & 0.60 & 0.85 & & 0.66 \\
HSM~\cite{yang2019hierarchical} & 0.55 & \textbf{0.63} & 0.98  & 0.62 & 0.49 & 0.31 & 0.04 & 0.60 & 0.64 & 0.83 & 0.18 & 0.87 & 0.23 & 0.70 & \textbf{0.90} & & 0.57 \\
Ours & \textbf{0.87} & 0.60 & \textbf{0.99} & \textbf{0.78} & \textbf{0.86} & \textbf{0.55} & 0.61 & \textbf{0.75} & 0.75 & \textbf{0.98} & \textbf{0.64} & \textbf{0.95} & \textbf{0.79} & \textbf{0.81} & 0.88 & & \textbf{0.79} \\
\bottomrule
\end{tabular} \hspace{1mm}
\begin{tabular}{cc}
\toprule
  Avg bad-4.0 \\ 
\midrule
  17.57 \\ 
  20.68 \\
  17.58 \\
  \textbf{5.92} \\
  16.07\\
\bottomrule
\end{tabular}
\includegraphics[width=0.9\textwidth]{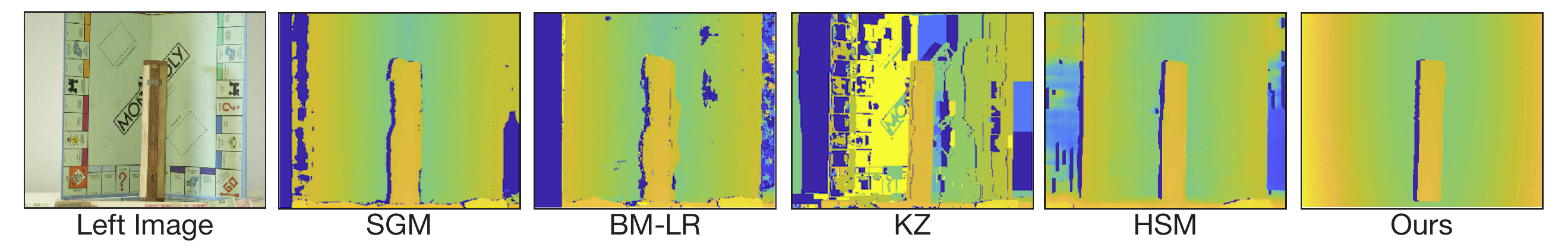}
\vspace{-3mm}
\caption{Top: Accuracy of occlusion boundaries measured by occlusion F1-score in fifteen scenes. Rightmost column is bad-4.0 disparity error. Bottom: Our final disparity/occlusion map compared to others for Scene 1 (occlusions in dark blue).}
\label{tbl:result}
\vspace{-2mm}
\end{table*}

\vspace{1mm}
\noindent \textbf{Updating global shapes and boundary.}
\label{sec:global_updates}
The global models $\mathbf{{\Theta}}_{j=1,2}$ are updated by maximum likelihood estimation:
\begin{equation}
\small
\begin{split}
    \mathbf{\Theta}_{j=1,2} & =  \argmin\limits_{\mathbf{\Theta}_j} \sum_{\substack{(x,y) \in \Omega_{j}}}
    \tfrac{(\Theta_j(x,y) - \overline{d}(x,y))^{2}}{2 \sigma^{2}(x,y)},
\end{split}
\end{equation}
with $\Omega_1,\Omega_2$ the foreground and background respectively. This requires solving two linear systems of equations based on the consensus values at many pixels. 
The last row of Fig.~\ref{fig:patch_and_pixel_vote}(b) shows an example. The erroneous consensus means are successfully down-weighted by $1/\sigma(x,y)$, thus the global shapes are correct.
Finally, the function $\phi(x,y)$ is updated using gradient descent with a discrete approximation to Eqn.~\ref{eqn:dphi_dt}.

 \begin{figure}[t]
     \centering
     \includegraphics[width=0.49\textwidth]{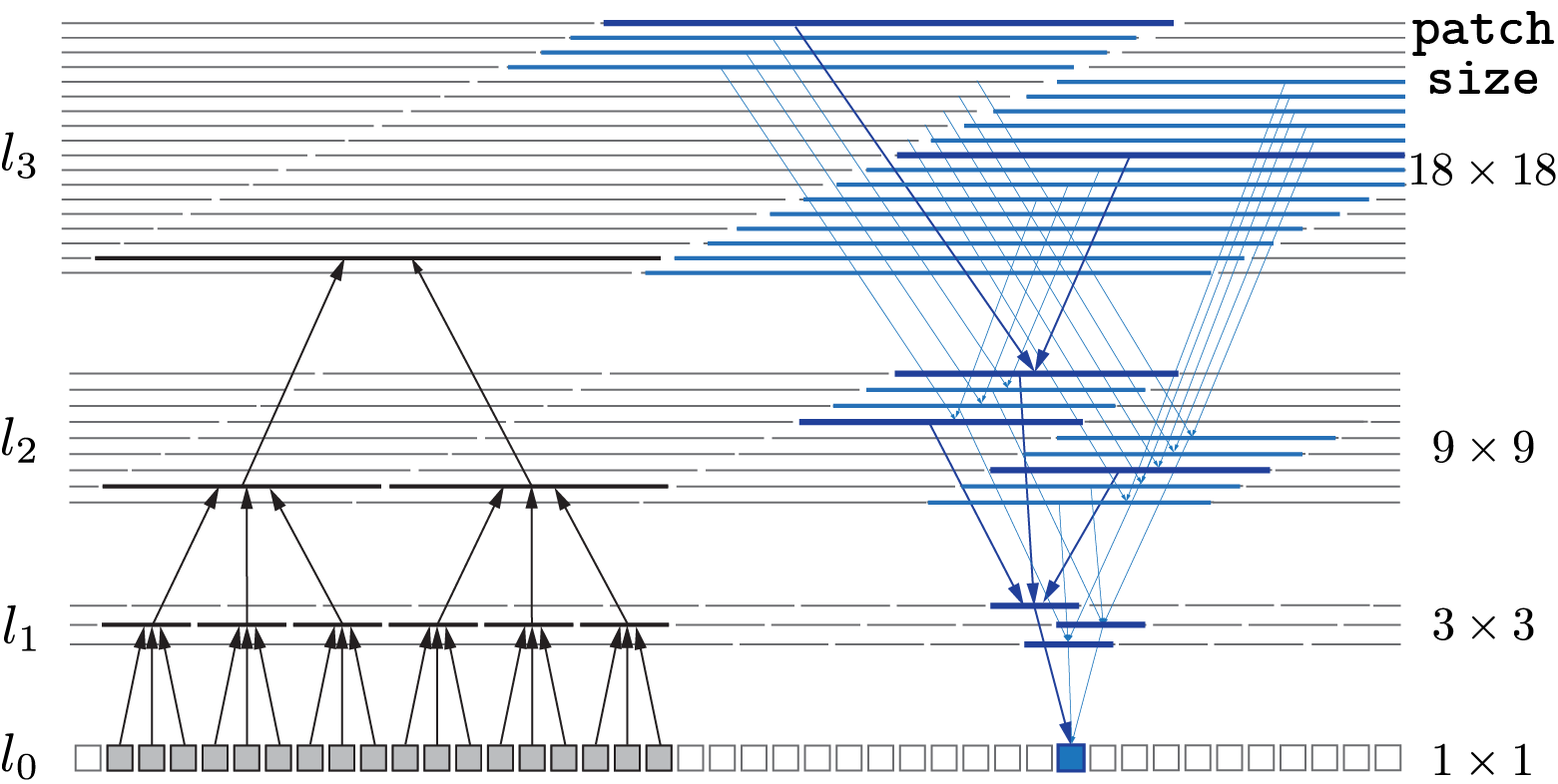}
    \caption{One-dimensional depiction of overlapping multi-scale patches, adapted from Chakrabarti et al.~\cite{chakrabarti2015low}. Layer $l_0$ consists of single pixels, and larger patches are composed of non-overlapping smaller ones.}
         \label{fig:overlapping_patches}
\end{figure}

The above alternating updates can be implemented cooperatively using an undecimated, hierarchical set of patches at multiple scales, similar to that of Chakrabarti et al.~\cite{chakrabarti2015low}. As depicted in one dimension in Fig.~\ref{fig:overlapping_patches}, patches at level $\ell_0$ comprise single pixels, and patches at higher levels $\ell_i$ are unions of patches in levels below. Updates can occur via information that passes predominantly between parents and children in the hierarchy. Per-patch disparity messages can be updated in an upward pass using information from pixels each patch contains, with information being relayed from children to parents similar to ~\cite{chakrabarti2015low}. The consensus update can occur in a downward pass with each pixel aggregating information relayed from all of the patches that contain it. Finally, global shapes can be updated by cooperatively solving the linear systems using a hierarchical variant of consensus averaging~\cite{tron2011distributed}. 

\section{Experiments}
There are few stereo benchmarks with strictly bi-layer scenes, so we curate a small one using four photographic crops from Middlebury 2006~\cite{hirschmuller2007evaluation} and eleven renderings from Falling Things~\cite{tremblay2018falling}. The latter includes ground-truth foreground boundaries, and we manually annotate them in the former.

Our model can use any underlying matching and boundary signals. We choose simplicity: absolute difference of intensity for $M(x,y,d)$; inverse of epipolar Gaussian derivative ($\approx | \frac{\partial M(x,y,d)}{\partial x} | $) for $B_{\text{occ}}(x,y,d)$ (see~\cite{wang2017toward}); and for $B_{\text{mono}}(x,y,d)$, the inverse of the sum of Sobel filter responses in left and right projections of $(x,y,d)$. 
For each input, we manually initialize to an elliptical boundary. We re-initialize $\phi(x,y)$ to a signed distance function every 10 iterations, and we apply a $7\times7$ median filter to $\phi(x,y)$ after every iteration to eliminate thin structures. Parameters are: 
$dt= 0.2$; $\alpha_1= 0.2$; $\alpha_2= 0.8$; $\alpha_3= 0.1$; $\mu=4.0$ and $\beta= 0.4/d_\text{max}$

We compare to a variety of methods that estimate binary occlusion labels: SGM with occlusion label~\cite{hirschmuller2009matching}, block matching with LR-consistency check (BM-LR) (see e.g.,~\cite{fua1993parallel}), graph cuts with occlusion (KZ)~\cite{kolmogorov2001computing}, and a deep model HSM~\cite{yang2019hierarchical} that augments training with fake occlusions. 
(We use the HSM authors' weights, trained on a mixture of datasets including Middlebury.)
We measure boundary accuracy using precision and recall---summarized by F1 score---of estimated binary occlusion labels in a region of $\pm 20$ epipolar pixels around the ground-truth boundary. As a secondary comparison, we measure accuracy of estimated disparities in the true, mutually-visible subset of this region, using the fraction of pixels at which the disparity error is greater than four pixels (bad-4.0).


\begin{figure}[h!]
    \centering
    \includegraphics[width=0.46\textwidth]{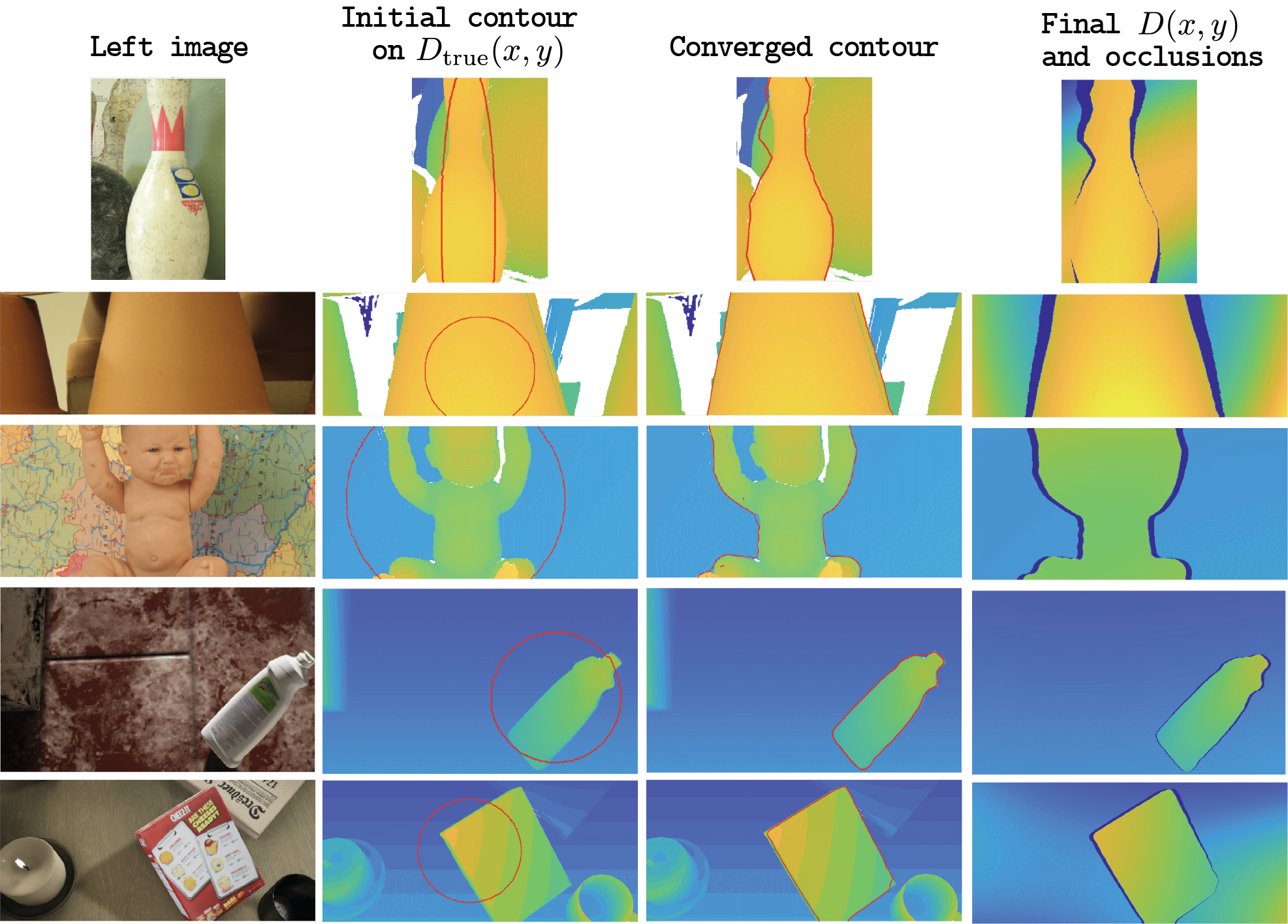}
    \vspace{-2mm}
    \caption{Results of scenes 2-6: left images, initial and converged boundary superimposed on true disparity maps, and the final $D(x,y)$. White pixels are holes in the ground truth.}
    \vspace{-2mm}
    \label{fig:more_viz}
\end{figure}

The results are summarized in Table~\ref{tbl:result}. Our model provides the highest occlusion F1-score in all but four scenes, and its disparity error near the boundaries (bad-4.0 error) is better than most. The deep model, HSM, achieves substantially lower disparity error, but its occlusion accuracy is middling. This supports the belief that iterative bottom-up models like ours could complement deep feed-forward models. 
Figure~\ref{fig:more_viz} visualizes the initialization and final results of our model, superimposed on the ground-truth disparity maps. Comparing this ground-truth disparity to our model's prediction $D(x,y)$ highlights a property of our model's disparity errors: they are predominantly caused by the crude quadratic approximation of foreground and background shape. These types of errors could be improved, for example, by executing a second, higher-fidelity stage of stereo processing in each region separately (i.e., ``surface plus parallax''). 


\section{Discussion}

In summary, our algorithm can be used as-is for scenes with isolated foreground objects and for tasks, like grasping, that rely heavily on having an accurate foreground mask. Broadening to general scenes and tasks will require upgrading the underlying matching and boundary signals to more sophisticated (e.g.,~``deep''~\cite{vzbontar2016stereo,wang2019local,xie2015holistically}) alternatives, and perhaps tuning the parameters jointly by unrolling our model's iterations in time. Open question include how to increase the number of regions beyond two (e.g., using multi-phase level sets~\cite{vese2002multiphase}), and how to combine this with complementary top-down information from deep feed-forward models.
\bibliographystyle{IEEEbib}
\bibliography{refs}

\begin{thebibliography}{10}

\bibitem{poggi2020synergies}
Matteo Poggi, Fabio Tosi, Konstantinos Batsos, Philippos Mordohai, and Stefano
  Mattoccia,
\newblock ``On the synergies between machine learning and stereo: a survey,''
\newblock {\em arXiv:2004.08566}, 2020.

\bibitem{wang2019local}
Jialiang Wang and Todd Zickler,
\newblock ``Local detection of stereo occlusion boundaries,''
\newblock in {\em Computer Vision and Pattern Recognition}, 2019.

\bibitem{middlebury_website}
``The middlebury stereo benchmark website,''
\newblock in {\em
  \href{http://vision.middlebury.edu/stereo/}{http://vision.middlebury.edu/stereo/}}.

\bibitem{belhumeur1996bayesian}
Peter~N Belhumeur,
\newblock ``A bayesian approach to binocular steropsis,''
\newblock in {\em International Journal of Computer Vision}, 1996.

\bibitem{bobick1999large}
Aaron~F Bobick and Stephen~S Intille,
\newblock ``Large occlusion stereo,''
\newblock {\em International Journal of Computer Vision}, 1999.

\bibitem{wang2017toward}
Jialiang Wang, Daniel Glasner, and Todd Zickler,
\newblock ``Toward perceptually-consistent stereo: A scanline study,''
\newblock in {\em International Conference on Computer Vision}, 2017.

\bibitem{zitnick2000cooperative}
C~Lawrence Zitnick and Takeo Kanade,
\newblock ``A cooperative algorithm for stereo matching and occlusion
  detection,''
\newblock {\em Transactions on Pattern Analysis and Machine Intelligence},
  2000.

\bibitem{kolmogorov2001computing}
Vladimir Kolmogorov and Ramin Zabih,
\newblock ``Computing visual correspondence with occlusions using graph cuts,''
\newblock in {\em International Conference on Computer Vision}, 2001.

\bibitem{lin2004surfaces}
MH~Lin and C~Tomasi,
\newblock ``Surfaces with occlusions from layered stereo.,''
\newblock {\em Transactions on Pattern Analysis and Machine Intelligence},
  2004.

\bibitem{yamaguchi2012continuous}
Koichiro Yamaguchi, Tamir Hazan, David McAllester, and Raquel Urtasun,
\newblock ``Continuous markov random fields for robust stereo estimation,''
\newblock in {\em European Conference on Computer Vision}, 2012.

\bibitem{fua1993parallel}
Pascal Fua,
\newblock ``A parallel stereo algorithm that produces dense depth maps and
  preserves image features,''
\newblock {\em Machine vision and applications}, 1993.

\bibitem{sun2005symmetric}
Jian Sun, Yin Li, Sing~Bing Kang, and Heung-Yeung Shum,
\newblock ``Symmetric stereo matching for occlusion handling,''
\newblock in {\em Computer Vision and Pattern Recognition}, 2005.

\bibitem{hirschmuller2007stereo}
Heiko Hirschmuller,
\newblock ``Stereo processing by semiglobal matching and mutual information,''
\newblock in {\em Transactions on Pattern Analysis and Machine Intelligence},
  2007.

\bibitem{yang2019hierarchical}
Gengshan Yang, Joshua Manela, Michael Happold, and Deva Ramanan,
\newblock ``Hierarchical deep stereo matching on high-resolution images,''
\newblock in {\em Computer Vision and Pattern Recognition}, 2019.

\bibitem{chakrabarti2015low}
Ayan Chakrabarti, Ying Xiong, Steven~J Gortler, and Todd Zickler,
\newblock ``Low-level vision by consensus in a spatial hierarchy of regions,''
\newblock in {\em Computer Vision and Pattern Recognition}, 2015.

\bibitem{caselles1997geodesic}
Vicent Caselles, Ron Kimmel, and Guillermo Sapiro,
\newblock ``Geodesic active contours,''
\newblock in {\em International Journal of Computer Vision}. 1997, Springer.

\bibitem{chan2001active}
Tony~F Chan and Luminita~A Vese,
\newblock ``Active contours without edges,''
\newblock in {\em Transactions on Image Processing}, 2001.

\bibitem{hirschmuller2009matching}
H~Hirschmuller and D~Scharstein,
\newblock ``Evaluation of stereo matching costs on images with radiometric
  differences,''
\newblock in {\em Transactions on Pattern Analysis and Machine Intelligence},
  2009.

\bibitem{tron2011distributed}
Roberto Tron and Ren{\'e} Vidal,
\newblock ``Distributed computer vision algorithms through distributed
  averaging,''
\newblock in {\em Computer Vision and Pattern Recognition}, 2011.

\bibitem{hirschmuller2007evaluation}
Heiko Hirschmuller and Daniel Scharstein,
\newblock ``Evaluation of cost functions for stereo matching,''
\newblock in {\em Computer Vision and Pattern Recognition}, 2007.

\bibitem{tremblay2018falling}
Jonathan Tremblay, Thang To, and Stan Birchfield,
\newblock ``Falling things: A synthetic dataset for {3D} object detection and
  pose estimation,''
\newblock in {\em Computer Vision and Pattern Recognition Workshops}, 2018.

\bibitem{vzbontar2016stereo}
Jure {\v{Z}}bontar and Yann LeCun,
\newblock ``Stereo matching by training a convolutional neural network to
  compare image patches,''
\newblock in {\em Journal of Machine Learning Research}, 2016.

\bibitem{xie2015holistically}
Saining Xie and Zhuowen Tu,
\newblock ``Holistically-nested edge detection,''
\newblock in {\em International Conference on Computer Vision}, 2015.

\bibitem{vese2002multiphase}
Luminita~A Vese and Tony~F Chan,
\newblock ``A multiphase level set framework for image segmentation using the
  {Mumford} and {Shah} model,''
\newblock in {\em International Journal of Computer Vision}, 2002.

\end{thebibliography}

\end{document}